\documentclass[conference]{IEEEtran}
\IEEEoverridecommandlockouts
\usepackage{cite}
\usepackage{amsmath,amssymb,amsfonts}
\usepackage{algorithmic}
\usepackage{graphicx}
\usepackage{textcomp}
\usepackage{listings}
\usepackage{xcolor}
\usepackage{bm}
\def\BibTeX{{\rm B\kern-.05em{\sc i\kern-.025em b}\kern-.08em
    T\kern-.1667em\lower.7ex\hbox{E}\kern-.125emX}}
\begin{document}

% \title{Simulating an Autonomous System in Carla using ROS2\\[-0.3em]
% {\Large Atlas Racing - Heriot Watt University Dubai}\\[-0.6em]
% }

% \author{\IEEEauthorblockN{Aditya Shibu, Joseph Abdo, Moaiz Saeed, Apsara Sivaprazad, Abdul Maajid Aga, and Dr Mohammed Al Musleh}
% }

\title{Simulating an Autonomous System in CARLA using ROS 2\\
}

\makeatletter
\newcommand{\linebreakand}{%
  \end{@IEEEauthorhalign}
  \hfill\mbox{}\par
  \mbox{}\hfill\begin{@IEEEauthorhalign}
}
\makeatother

\author{%
\IEEEauthorblockN{Joseph Abdo}
\IEEEauthorblockA{\textit{Engineering and Physical Sciences}\\
\textit{Heriot Watt University}\\ Dubai, UAE\\ jwa2001@hw.ac.uk}
\and
\IEEEauthorblockN{Aditya Shibu}
\IEEEauthorblockA{\textit{Mathematical and Computer Sciences}\\
\textit{Heriot Watt University}\\ Dubai, UAE\\ as2397@hw.ac.uk}
\and
\IEEEauthorblockN{Moaiz Saeed}
\IEEEauthorblockA{\textit{Engineering and Physical Sciences}\\
\textit{Heriot Watt University}\\ Dubai, UAE\\ ms2277@hw.ac.uk}
\linebreakand
\IEEEauthorblockN{Abdul Maajid Aga}
\IEEEauthorblockA{\textit{Engineering and Physical Sciences}\\
\textit{Heriot Watt University}\\ Dubai, UAE\\ ama2024@hw.ac.uk}
\and
\IEEEauthorblockN{Apsara Sivaprazad}
\IEEEauthorblockA{\textit{Engineering and Physical Sciences}\\
\textit{Heriot Watt University}\\ Dubai, UAE\\ as2404@hw.ac.uk}
\and
\IEEEauthorblockN{Dr. Mohamed Al Musleh}
\IEEEauthorblockA{\textit{School of Textiles and Design}\\
\textit{Heriot Watt University}\\ Dubai, UAE\\ m.al-musleh@hw.ac.uk}
}

\maketitle

\begin{abstract}
Autonomous racing offers a rigorous setting to stress test perception, planning, and control under high speed and uncertainty. This paper proposes an approach to design and evaluate a software stack for an autonomous race car in CARLA: Car Learning to Act simulator, targeting competitive driving performance in the Formula Student UK Driverless (FS-AI) 2025 competition. By utilizing a 360° light detection and ranging (LiDAR), stereo camera, global navigation satellite system (GNSS), and inertial measurement unit (IMU) sensor via ROS 2 (Robot Operating System), the system reliably detects the cones marking the track boundaries at distances of up to 35 m. Optimized trajectories are computed considering vehicle dynamics and simulated environmental factors such as visibility and lighting to navigate the track efficiently. The complete autonomous stack is implemented in ROS 2 and validated extensively in CARLA on a dedicated vehicle (ADS-DV) before being ported to the actual hardware, which includes the Jetson AGX Orin 64GB, ZED2i Stereo Camera, Robosense Helios 16P LiDAR, and CHCNAV Inertial Navigation System (INS).
\end{abstract}


\section{Literature Review}

The Formula Student Driverless (FS-AI) competition has stimulated research on autonomous racing software stacks validated through both real world testing and simulation. Prior work has demonstrated that simulation is critical in enabling iterative testing, reducing cost, and mitigating risks before deployment on physical racecars.

Several platforms have been used to simulate autonomous racing. Culley et al. \cite{b19} developed their driverless system within Gazebo, combining stereo vision with YOLO(You Only Look Once)-based cone detection and an Extended Kalman Filter (EKF) for localization \cite{Kou_2023}. While their framework enabled tuning of controllers and resulted in measurable lap time improvements, the limitations of Gazebo, including simplified dynamics and lack of high fidelity sensor models, constrained the ability to validate the entire perception and control pipeline. Kabzan et al. \cite{kabzan2019amz} presented the AMZ Driverless architecture, which combined redundant LiDAR and vision based cone detection pipelines, FastSLAM 2.0 for mapping sparse cone environments, and Model Predictive Control (MPC) for near-optimal vehicle handling.

Across these works, perception and mapping remain central challenges. Stereo vision systems, as employed by Culley et al. \cite{b19}, faced reduced reliability for long range cone detection, while LiDAR-camera redundancy in Kabzan et al. \cite{kabzan2019amz} improved robustness but at the expense of greater system complexity and computational demand. In comparison, the present work employs a YOLOv8-based cone detector together with low level LiDAR–camera fusion. This approach provides higher accuracy and robustness than the YOLOv3-Tiny stereo vision pipeline in \cite{b19}, while avoiding the computational overhead of the redundant perception architecture in \cite{kabzan2019amz}. The fusion system further adapts confidence weighting dynamically according to sensor reliability, ensuring consistent performance across diverse conditions.

Localization methods have also differed. Culley et al. \cite{b19}  relied primarily on GPS(Global Positioning System)–IMU(Inertial Measurement Unit) fusion with an EKF, which degraded under GPS drift. In this work, GNSS(Global Navigation Satellite System)–IMU fusion is integrated with Google Cartographer and LiDAR based cone mapping, producing a more accurate and extensible Simultaneous Localization and Mapping (SLAM) framework while maintaining computational efficiency. This addresses the limitations of Culley et al. \cite{b19} approach by reducing dependency on GPS accuracy alone and enabling simultaneous online mapping of cone markers.

Path planning approaches in prior works also reflect trade offs. Heuristic midpoint methods proved effective for simple layouts but scaled poorly to complex track topologies, while uncertainty aware graph search techniques introduced by Kabzan et al. \cite{kabzan2019amz} offered robustness at the cost of runtime efficiency. Control strategies in autonomous racing similarly diverged. Simpler methods such as Pure Pursuit for lateral control and PID for longitudinal control, as applied by Culley et al. \cite{b19}, provide lightweight and interpretable implementations but struggle near vehicle handling limits. By contrast, predictive controllers such as MPC, demonstrated by Kabzan et al. \cite{kabzan2019amz}, directly optimize trajectories with respect to tire and track constraints but require significant computational resources to operate in real time. In comparison, the adaptive Pure Pursuit controller with velocity dependent lookahead used in this work provides a balanced alternative as it maintains the efficiency of geometric tracking while dynamically adjusting the look ahead distance based on track context, improving stability in straights and responsiveness in tight corners. This adaptability enables smoother path tracking than the fixed parameter Pure Pursuit of \cite{b19}, while avoiding the computational burden of MPC \cite{kabzan2019amz}.

Simulation has historically been limited by either low fidelity sensor modeling (Gazebo, FSSIM) or abstracted perception inputs that omit realistic sensing constraints. Similarly, perception and localization pipelines have either traded off long range accuracy for simplicity (stereo vision, EKF) or improved robustness through redundancy and particle filtering at the cost of increased system overhead. Control methods have oscillated between computationally efficient but suboptimal designs and performance-oriented but resource heavy predictive controllers. These gaps motivate the use of high fidelity simulation frameworks such as CARLA \cite{DBLP:journals/corr/abs-1711-03938}, coupled with modular middleware such as ROS 2 \cite{ros2humble}, to evaluate the full perception–planning–control pipeline under realistic sensing and dynamic conditions.

\section{Methodology}
The autonomous driving system is simulated in CARLA with a 3D LiDAR, Stereo Camera, GNSS, and IMU providing perception and localisation data via ROS 2. Cone markers are detected by fusing stereo imagery and LiDAR point clouds, with mapping achieved through a modified 3D Cartographer using GNSS–IMU fusion via an EKF. Path planning computes midpoints between cones, and control is executed through a Pure Pursuit controller for steering and a PID controller for longitudinal speed regulation.

\subsection{System Architecture}
This section outlines the topology of the autonomous system that was simulated, which includes sensors, processing, and control as shown in Fig.~\ref{fig1}. ROS 2 Humble is the core part of this system architecture, serving as the main inter-node communication framework that facilitates data exchange, synchronization, and modular integration of the various software components.

The system comprises of a 3D LiDAR, Stereo Camera, IMU, and GNSS, with ROS 2 Humble. Visualization tools such as ROS Visualization vr. 2 (RViz2) and Foxglove were extensively used to monitor sensor data, system states, and mapping outputs, facilitating easier diagnosis and iterative development throughout the simulation and testing process.

\begin{figure}[htbp]
\centerline{\includegraphics[scale=0.28]{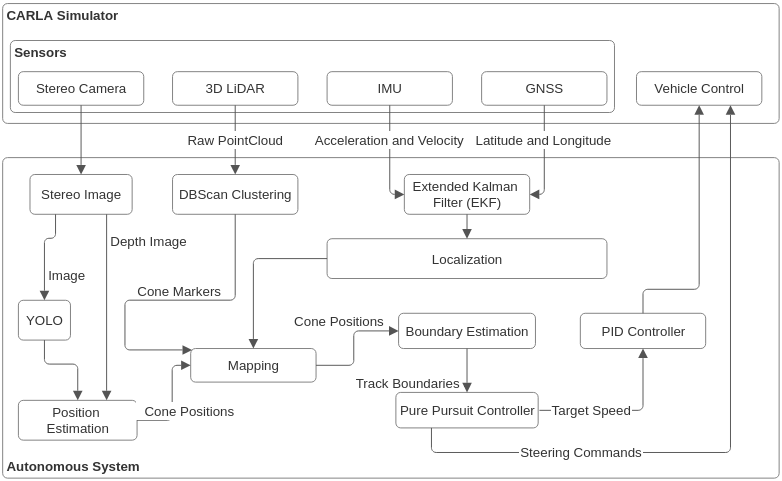}}
\caption{Integrated Sensor-to-Control Architecture for Autonomous Navigation.}
\label{fig1}
\end{figure}

\subsection{Perception}
Track boundaries are defined by coloured cones, making their accurate detection and localization essential for constructing a reliable map of the circuit. Our perception system leverages stereo camera data to capture colour imagery and depth information, which we fuse with cone detections published by the filtered LiDAR point cloud. By integrating these sensor inputs, the system identifies cones and estimates their positions relative to the vehicle in real time. This combined visual and spatial data enables an ego-centric representation of the track layout, forming the basis for planning and control within the simulation.

\subsubsection{Transform frames}
In our system, all sensors and processing modules operate within a unified tf2 tree in ROS 2, ensuring a common spatial reference for data fusion and state estimation. The primary frames include \textbf{map}, \textbf{odom}, \textbf{base\_link}, \textbf{imu\_link}, \textbf{lidar\_link}, and \textbf{gnss\_link}, as depicted in Fig.~\ref{fig2}.

\begin{figure}[htbp]
\centerline{\includegraphics[scale=0.45]{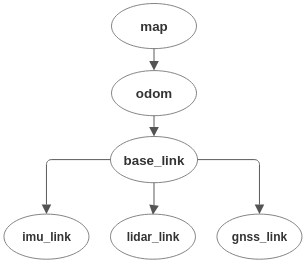}}
\caption{TF2 frame tree of the autonomous system}
\label{fig2}
\end{figure}

\subsubsection{LiDAR based Cone Perception}
The system simulates a 3D LiDAR sensor with a 16-beam configuration. To improve the quality and reliability of sensor data, filtering and clustering methods are applied to eliminate noise and irrelevant points that may negatively impact path planning. The LiDAR-based cone detection process is implemented through a multi-stage pipeline, described as follows:

\begin{itemize}
\item Raw Data Acquisition:

Data is captured as 3D points and transformed into the vehicle's reference frame using the corresponding sensor transforms. Specifically, the system utilizes the \textbf{lidar\_link} tf frame to accurately convert LiDAR point cloud
data into a consistent coordinate frame relative to the vehicle. This enables reliable spatial interpolation necessary for subsequent processing and real-time performance.

\item Height-based filtering:
\begin{equation}
\mathit{near\_ground\_mask} =
\left( z < h_{\text{ground}} + 0.5 \right) \,\land\,
\left( z > h_{\text{ground}} \right)
\end{equation}
where \textbf{‘z’} is the z-coordinate of the LiDAR points, 
and \textbf{$h_{\text{ground}}$} denotes the estimated ground height.

\item Cone Identification:

Cone identification is performed using DBSCAN(Density-Based Spatial Clustering of Applications with Noise) clustering with parameters eps=0.5 and min\_samples=5, striking a good balance between noise reduction and cluster detection. DBSCAN was chosen over alternatives like OPTICS(Ordering Points To Identify the Clustering Structure) due to its superior computational efficiency O(n log n) vs. O(n\textsuperscript{2}) \cite{b8}, making it well-suited for real-time processing. While OPTICS offers more flexibility, DBSCAN effectively filters noise and reliably clusters cones given their uniform size and spacing \cite{b9}.
\vspace{0.2em}

\item Cone Validation:

Cone detection is done through size validation, ensuring that each cluster matches the expected physical dimensions of a cone. This is followed by confidence scoring, which evaluates point density and other cluster characteristics. Additionally, multi-frame tracking is implemented to reduce false positives and enhance detection stability over time.
\end{itemize}

\subsection{Camera based Cone Perception}
The object detection system is developed to reliably recognize and classify track cones in real time, delivering accurate spatial data essential for navigation. After evaluating multiple methods, YOLOv8 was chosen as the core detection model due to its balance of speed and accuracy. This YOLO model was also used on the real ADS-DV platform to detect the track boundaries as seen in Fig.~\ref{RealADS-DV}

\begin{figure}[htbp]
\centerline{\includegraphics[scale=0.12]{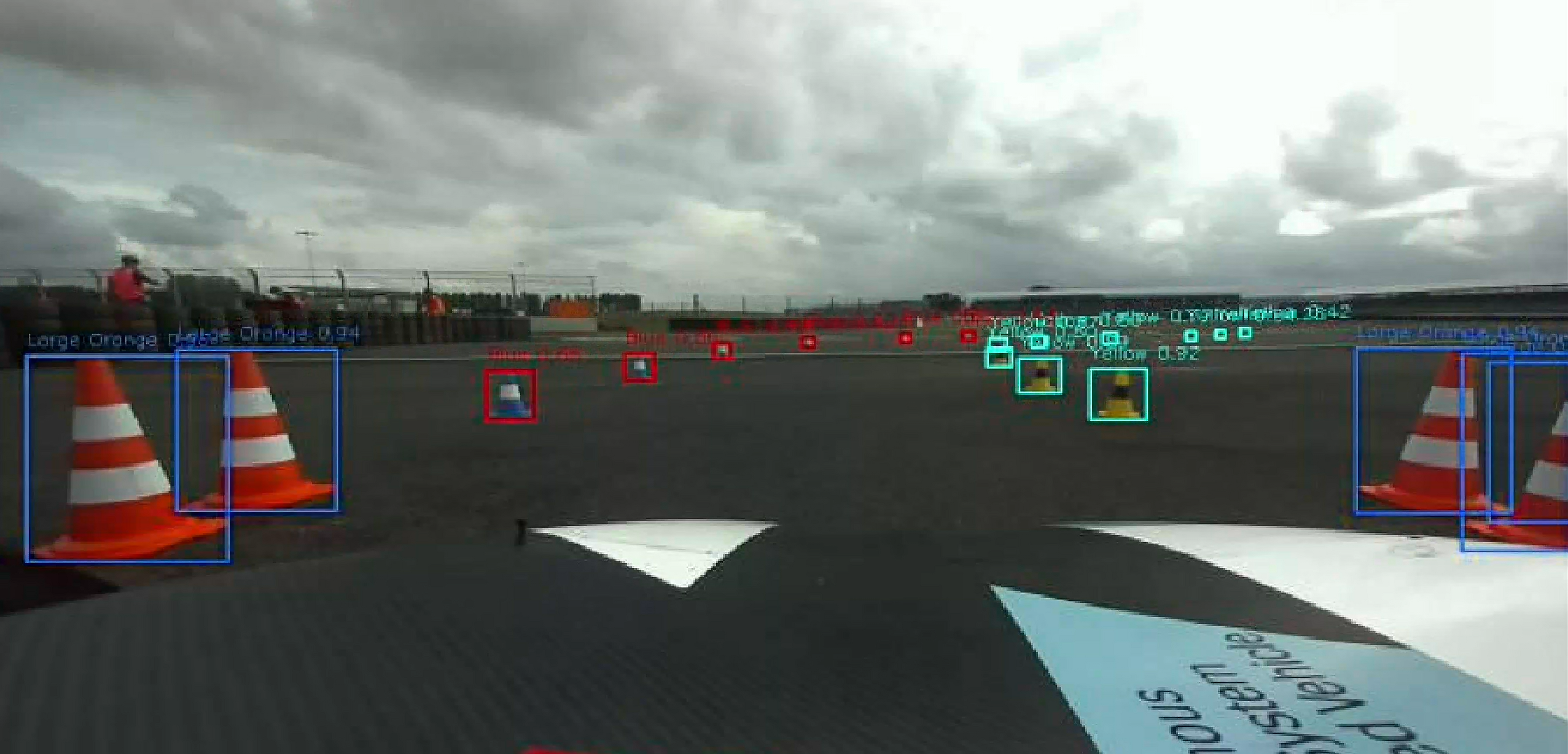}}
\caption{Real-world ADS-DV platform implementation utilizing the custom YOLOv8 model}
\label{RealADS-DV}
\end{figure}

\subsubsection{Model Development and Training}
Three cone datasets from Kaggle were created to form a unified training dataset with standardised labelling conventions. The classification scheme used numeric labels (0=orange, 1=yellow, 2=blue, 3=large orange, and 4=unknown) to ensure consistent identification across various racing environments. The training process involved a train-test split of 80\% (10,003 images) 20\% (10\% validation, and 10\% testing, 1,250 images each) along with a training configuration with 100 epochs, 32 batches, and 25 early stopping patience \cite{b10}\cite{b11}\cite{b12}. 

Additionally, an updated detection model was trained using the FOSCO dataset, which comprises of data collected from over 41 Formula Student Teams, containing over 11,500 images and more than 220,000 annotated cones.\cite{fsoco}.

\begin{figure}[htbp]
\centerline{\includegraphics[scale=0.23]{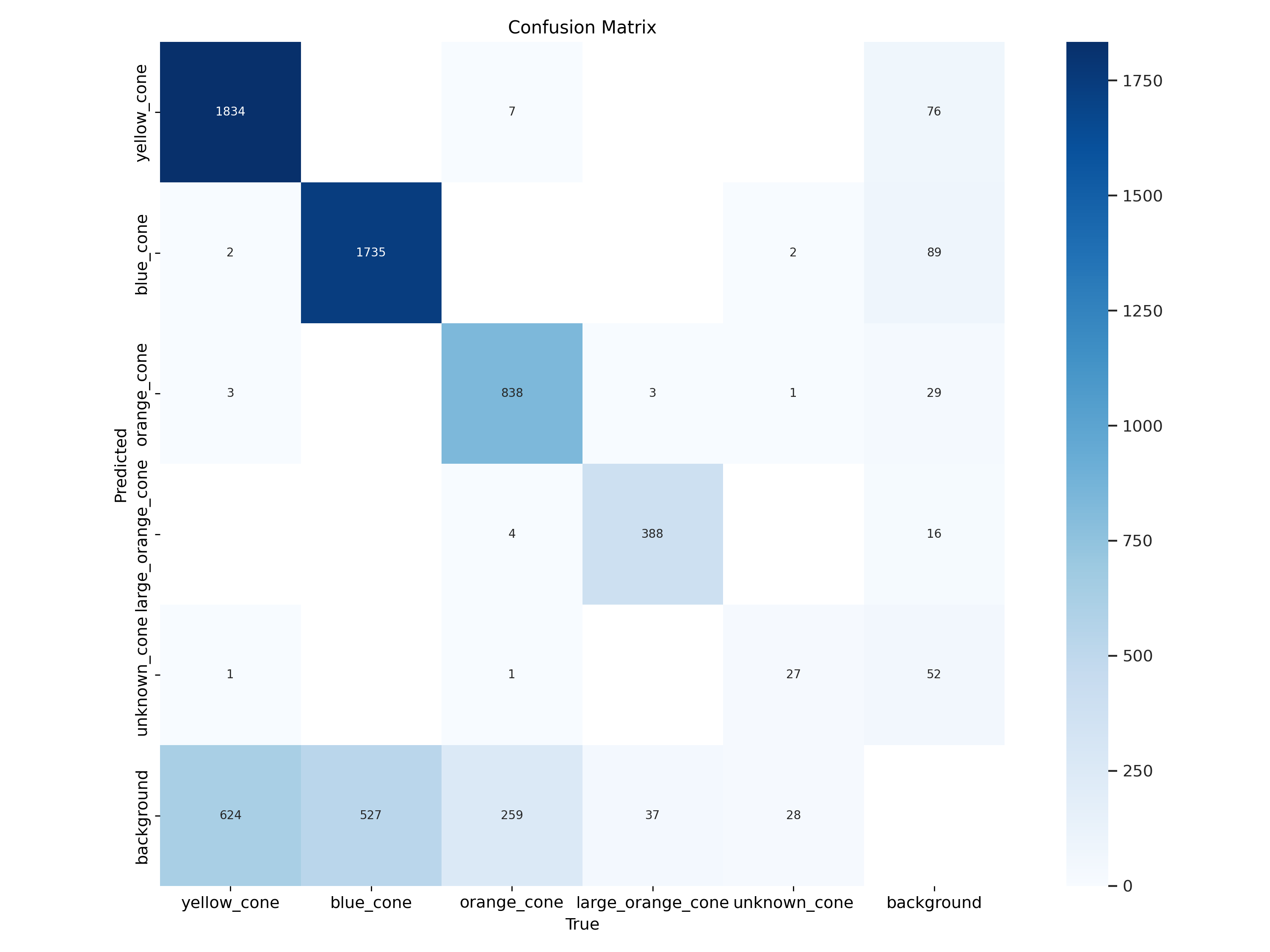}}
\caption{Confusion Matrix of the trained YOLOv8 Model}
\label{fig3}
\end{figure}

\textbf{Testing Result:} The model accuracy is shown in Fig.~\ref{fig3}, with appropriate results for yellow and blue cones (highest actual favourable rates). The confusion matrices revealed that most misclassifications occurred between visually similar cone types.

\subsubsection{Performance metrics and Evaluation}
Model evaluation demonstrated the relationship between confidence thresholds and detection accuracy. The F1 score peaked between confidence thresholds of 0.1 and 0.6. Precision remained high across most confidence levels for all cone classes. Recall started high at low confidence thresholds, stabilised between 0.6 and 0.4, and then dropped at higher thresholds.

\begin{figure}[htbp]
\centerline{\includegraphics[scale=0.3]{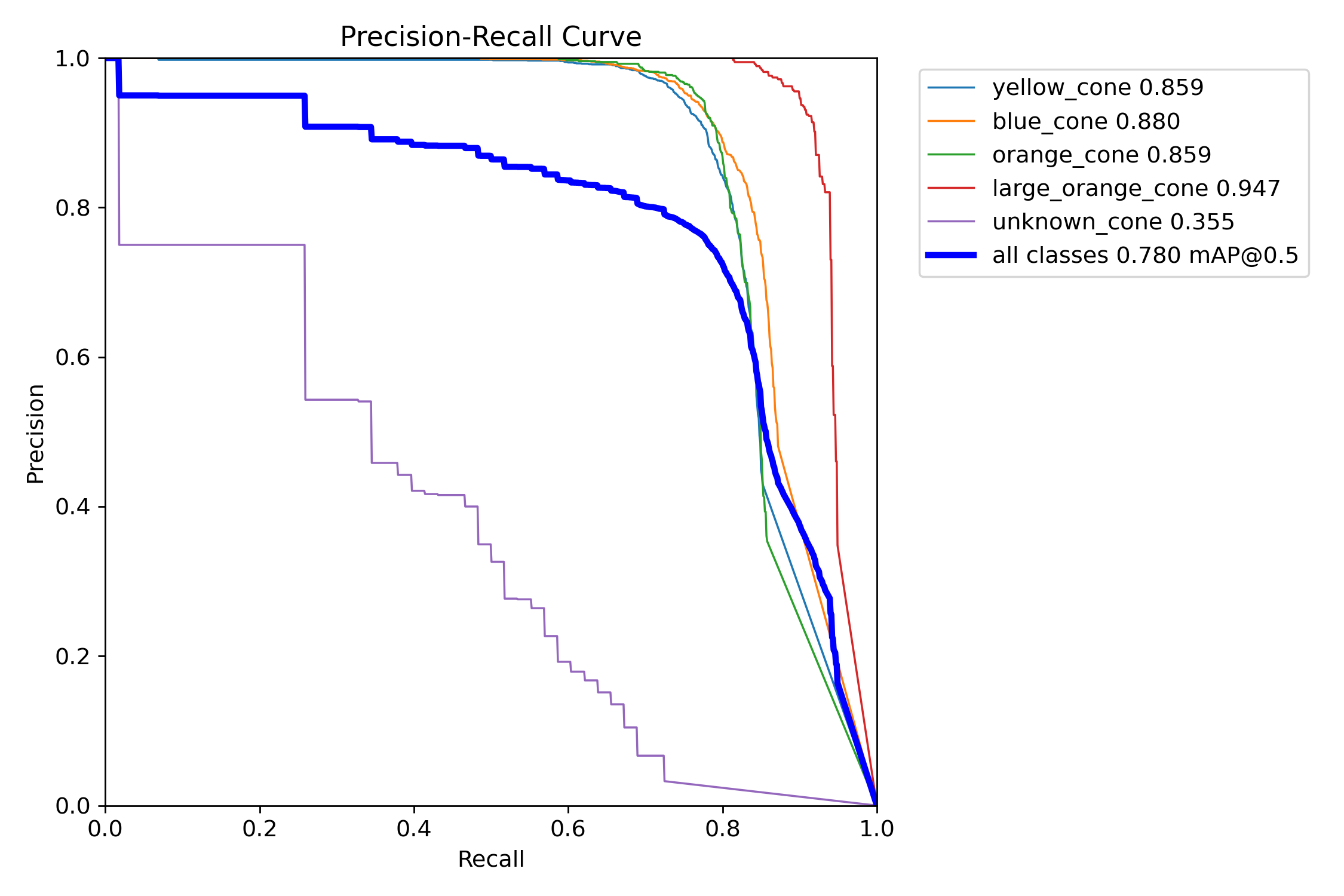}}
\caption{Precision-recall curve}
\label{fig4}
\end{figure}

\begin{figure}[htbp]
\centerline{\includegraphics[scale=0.25]{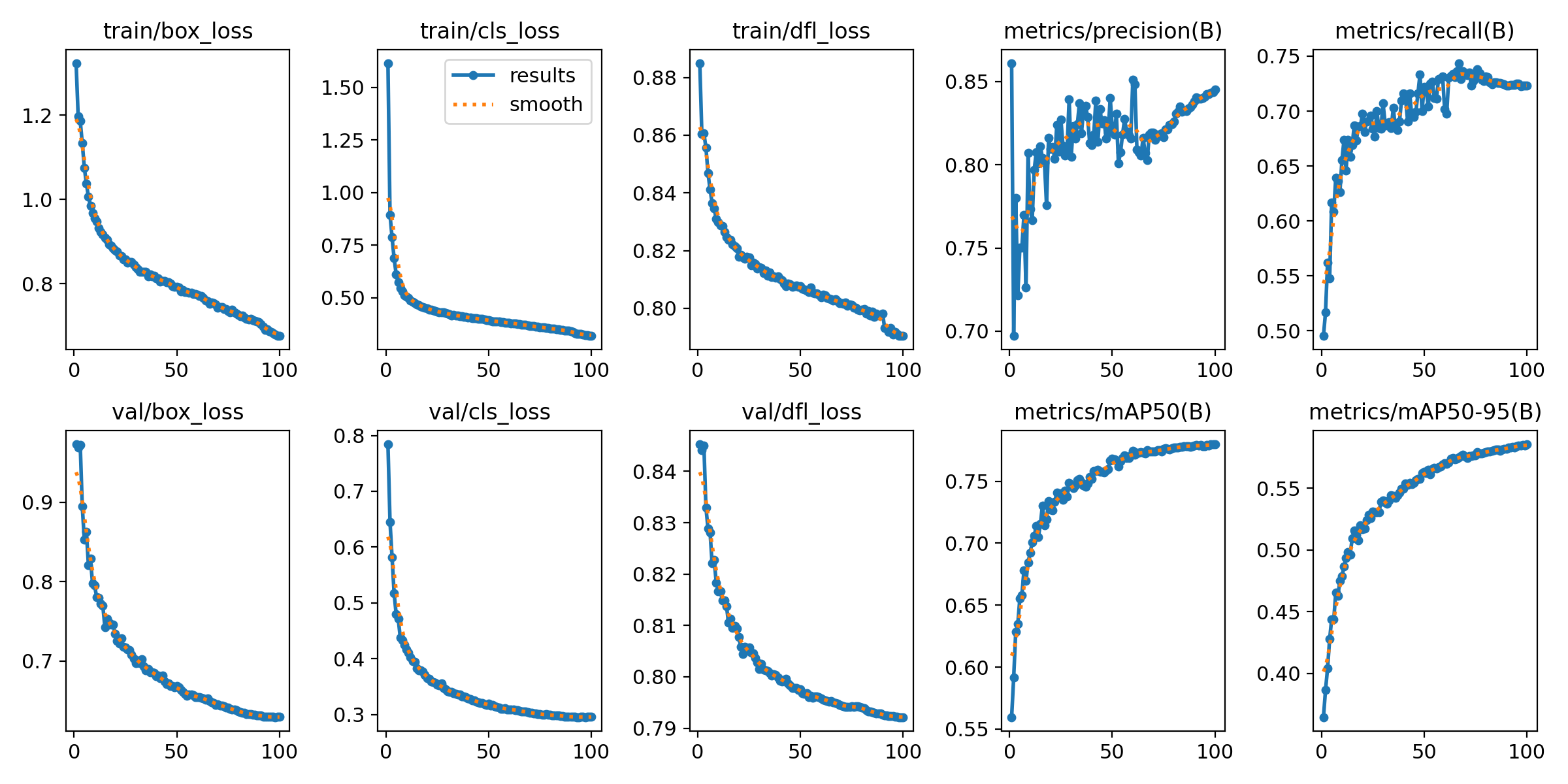}}
\caption{Training metrics graph}
\label{fig5}
\end{figure}

The precision-recall curve in Fig.~\ref{fig5} illustrates the trade-off between detection accuracy and coverage, with the final model achieving a \textbf{96\% cone detection accuracy}. The training metrics, as shown in Fig.~\ref{fig6}, demonstrated consistent improvement across epochs for bounding box loss, classification loss, and mean average precision (mAP), indicating stable training without significant overfitting.
Critically for real-time operation, the model achieves an average inference latency of \textbf{12--14 ms} on the simulation platform, enabling a high-frequency detection loop of approximately \textbf{71--83 Hz}.

\subsubsection{Depth Camera (vision-based)}
The system incorporates two simulated RGB cameras configured as a stereo pair to emulate the depth sensing capabilities of the ZED2i stereo camera. This simulated stereo setup is designed to replicate the essential features of the ZED2i, including providing depth estimation up to 20 meters, which is crucial for reliable cone localisation. By modelling the perception pipeline after the ZED2i, the simulated camera subsystem enhances system robustness by providing additional depth data from the car.

\begin{figure}[htbp]
\centerline{\includegraphics[scale=0.18]{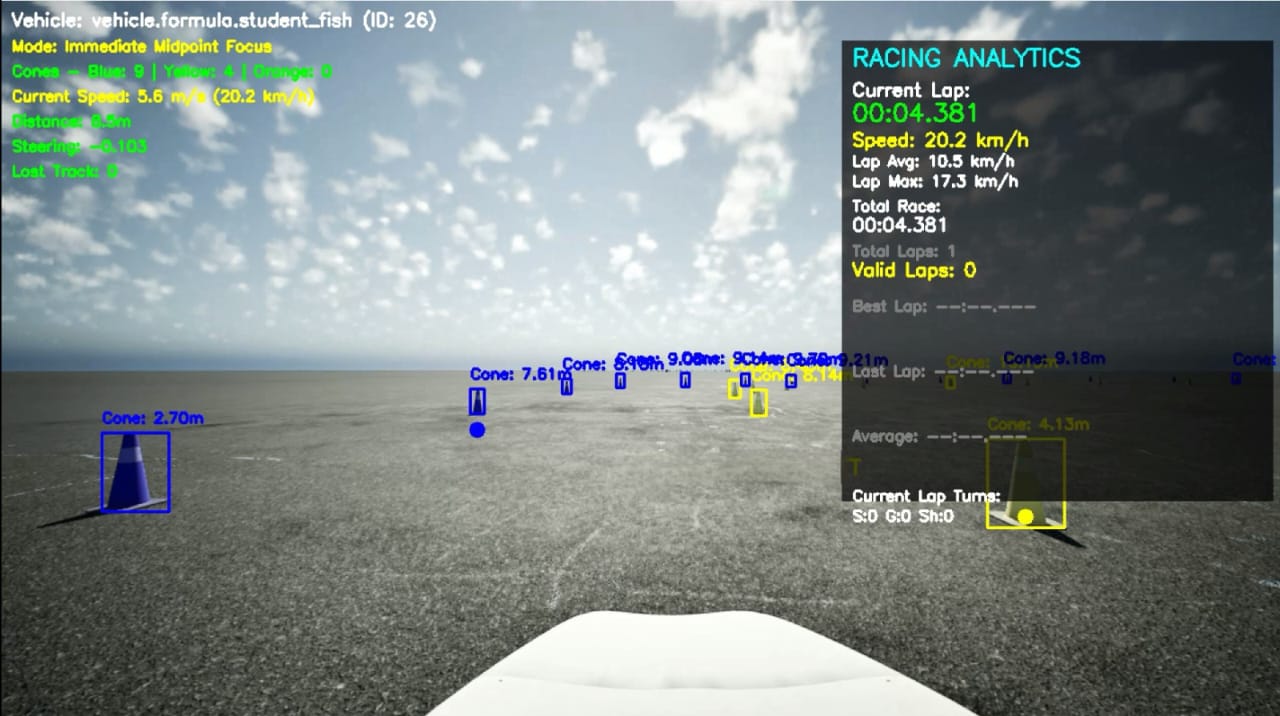}}
\caption{Simulator output displaying cone classification with distance estimated from depth data}
\label{simulator}
\end{figure}

\subsubsection{Advanced Depth Estimation}
A key innovation in the system is a multi-faceted depth estimation approach that combines several techniques: direct depth sampling from the stereo camera's depth map, temporal smoothing using a sliding window of recent detections, position-based depth estimation, and box–size–normalised depth correction. The final depth is computed as:
\begin{equation}
\textit{final\_depth} = 0.5 \cdot \textit{smoothed\_depth} + 0.5 \cdot \textit{position\_depth}
\end{equation}
This provides more reliable depth estimates than any single method alone, The depth estimations done by the system can be simulated as seen in Fig.~\ref{simulator}

\subsection{LiDAR-Camera sensor Fusion}\vspace{-0.4em}
The system employs a fusion technique to combine LiDAR data with camera detections, leveraging the strengths of both perception systems to create a more robust and reliable cone detection system.

The entire LiDAR processing pipeline, from raw point cloud acquisition through filtering and DBSCAN clustering, demonstrated an average processing time of \textbf{37.3 ms} (approx. \textbf{26.8 Hz}). This rate is sufficient for processing scans from the 16-beam LiDAR without significant data loss.

\subsubsection{Low-level Fusion Approach}
Our system implements low-level fusion to processes raw data from both sensors jointly before making detection decisions, providing key advantages such as enhanced detection accuracy through complementary sensor information, improved reliability across a wide range of weather conditions, more precise spatial localisation by combining depth information from multiple sources, and reduced false positives through cross-validation between sensors \cite{b14}.

\subsubsection{Sensor Alignment and Calibration}
The fusion pipeline begins with sensor alignment, utilising rigid transformation matrices to align the LiDAR and camera reference frames. It also incorporates depth correction factors to address systematic sensor biases and point cloud projection into the camera's image plane for correlation. These calibration steps ensure data from both sensors is registered in a standard coordinate frame before fusion.

\subsubsection{Data Association and Fusion Algorithm}
The core of the fusion system is a matching algorithm that associates LiDAR-detected cone clusters with camera-detected bounding boxes using the following steps:
\begin{itemize}
\item Projection Step: Lidar points are transformed to the camera frame and projected onto the image plane using the camera's intrinsic parameters
\item Association: Each 3D point is matched with its corresponding 2D detection using a weighted distance metric that considers both spatial proximity and confidence scores
\item Confidence Fusion: For matched detections, a confidence-weighted averaging is performed using the equation:
\begin{equation}
P_{\text{fused}} = \frac{P_{\text{camera}} \cdot w_{\text{camera}} + P_{\text{LiDAR}} \cdot w_{\text{LiDAR}}}{w_{\text{camera}} + w_{\text{LiDAR}}}
\end{equation}
Where \textbf{‘P'} stands for the position and \textbf{‘w'} stands for the weight/confidence \cite{b15}.
\item Classification Transfer: Colour information from the camera detections is transferred to the corresponding LiDAR points
\item Unmatched Point Handling: Points detected by only one sensor are preserved but assigned a reduced confidence score to maintain the continuity of detection
\end{itemize}

\subsubsection{Adaptive Confidence Weighting}
The system employs dynamic confidence weighting that adjusts according to sensor reliability under various conditions. For example, when Camera detection confidence is weighted higher in good lighting conditions, LiDAR measurements are prioritised in low-light environments or at longer distances. Distance-dependent weighting gives preference to the camera for colour classification and LiDAR for precise positioning, and Historical detection consistency is factored into confidence calculations.

This adaptive approach ensures optimal fusion performance across the diverse conditions encountered during the competition.

\vspace{-0.5em}

\subsection{Simultaneous localisation and mapping}

\vspace{-0.2em}
Our system leverages data from the GNSS, IMU, and 3D LiDAR to estimate the vehicle's state and construct a precise map of the track.

GNSS data is received as global-position messages in latitude and longitude, which are not directly usable in Cartesian coordinates. To convert these readings into a local Cartesian frame suitable for downstream processing, the system employs the \texttt{navsat\_transform} node. This node uses real-time orientation information from the IMU and reference position to transform GNSS \/fix data into a local odometry estimate within the map frame \cite{b17}

This GNSS-derived odometry is then fused with IMU measurement in an Extended Kalman Filter (EKF), producing a drift-corrected, centimetre-accurate pose estimate of the vehicle in the \texttt{odom} frame.

Mapping and localisation are achieved using Google's Cartographer, which subscribes to the filtered odometry from the EKF and to the 3D LiDAR scans. The LiDAR data is then passed through a customized DBScan algorithm to publish cone markers as the vehicle moves along, as seen in a 3D environment as seen in Fig.~\ref{Cone Markers}

This tightly coupled approach combines local odometry corrections with global GNSS constraints, maintaining accurate localisation without requiring a tightly integrated GNSS/IMU fusion. A high level architecture of our SLAM system can be seen in Fig.~\ref{SLAM Arch}

In performance testing, the EKF pose estimate updated at a high frequency (e.g., 100 Hz), while the Cartographer-based map and global localization loop achieved an update rate of \textbf{9.5 Hz (a 105 ms cycle time)}. This processing-intensive task was the primary system bottleneck, limited by the available computational resources of the simulation platform.

\begin{figure}[htbp]
\centerline{\includegraphics[scale=0.14]{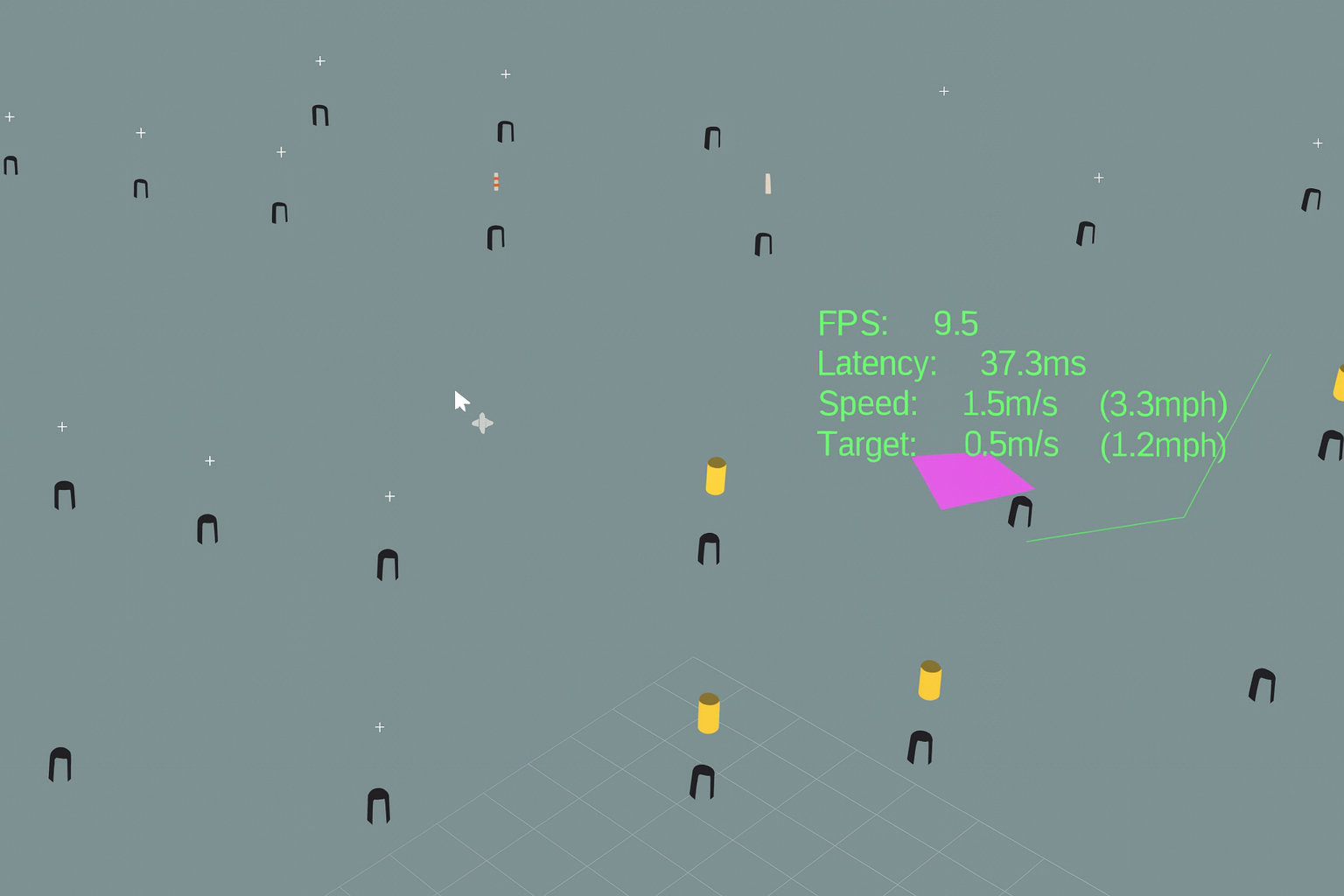}}
\caption{3D View of the cone markers placed by our SLAM algorithm}
\label{Cone Markers}
\end{figure}

%This tightly coupled approach combines local odometry corrections with global GNSS constraints, maintaining accurate localisation without requiring a tightly integrated GNSS/IMU fusion. 
Fig.~\ref{SLAM Arch} outlines our SLAM pipeline: GNSS/IMU are transformed and EKF-fused to yield \texttt{/odometry/filtered}, which seeds Cartographer~3D to fuse LiDAR scans into a global map for planning; Cartographer publishes \texttt{map}$\to$\texttt{odom} while the EKF provides \texttt{odom}$\to$\texttt{base\_link}, avoiding pose jumps.

\begin{figure}[htbp]
\centerline{\includegraphics[scale=0.15]{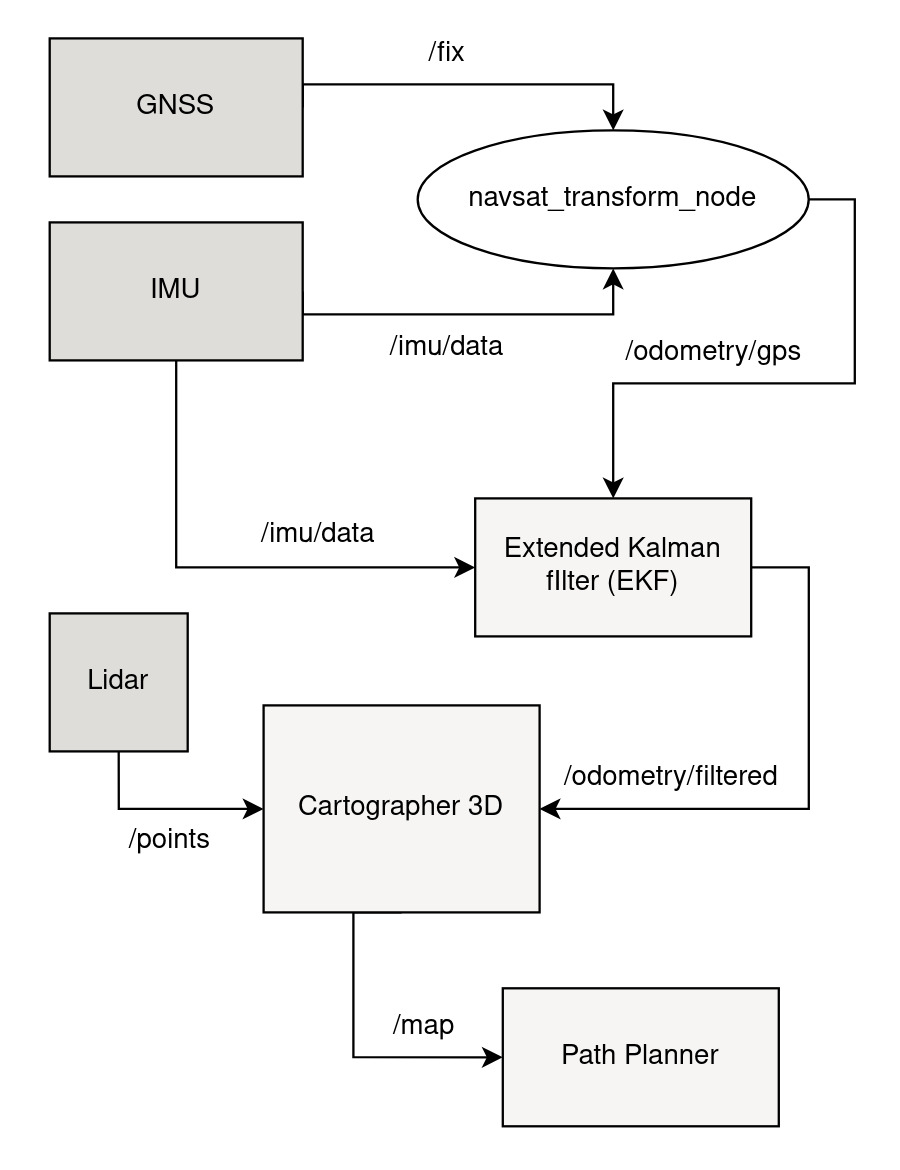}}
\caption{High-level SLAM Architecture}
\label{SLAM Arch}
\end{figure}

\subsection{Track Boundary Estimation}
Track boundary estimation was initially implemented using a nearest-neighbour heuristic to connect cones of the same colour, approximating the straightest line similar to the travelling salesman problem. While effective in simple layouts, this approach was computationally expensive and unreliable in complex scenarios such as junctions or overlapping lanes. It was therefore replaced with a simpler cone detection strategy, which reduces latency and improves responsiveness, while providing reliable input for the Pure Pursuit controller in path planning.

\subsection{Path Planning}
For path planning, the pure pursuit tracking algorithm \cite{b16} was employed to compute the steering commands required for the vehicle to follow the desired path in real-time. The method works by continuously selecting a goal point on the planned path ahead of the vehicle's current position at some lookahead distance \textbf{L\textsubscript{d}}. The algorithm features a lookahead distance of 2.5-5m, which is adaptively adjusted based on the car's current velocity to improve tracking stability according to the path's curvature. The lookahead distance is calculated by:
\begin{equation}
L_d = K_{ld} \times V + L_{fc}
\end{equation}

Where \textbf{‘L\textsubscript{d}'} is the dynamic lookahead distance, \textbf{‘L\textsubscript{ld}'} is the speed gain constant, \textbf{‘V'} is the instantaneous forward velocity of the vehicle, and \textbf{‘L\textsubscript{fc}'} is the minimum lookahead distance. 

Once the goal point is identified, the algorithm determines the required steering angle by using the formula:
\begin{equation}
\delta = \arctan\left(\frac{2L \sin(\alpha)}{L_d}\right)
\end{equation}

Where \textbf{‘L'} is the wheelbase of the vehicle, $\alpha$ is the angular deviation between the vehicle's heading and the vector connecting the real axle to the lookahead point, with \textbf{‘L\textsubscript{d}'} being the lookahead distance.

The calculation is repeated for each control cycle, ensuring the vehicle constantly aligns with the updated path based on the latest cone detections. Unlike map-based navigation systems, this implementation operates in a dynamic and unstructured environment where the vehicle relies solely on visual cone detection to build a local track representation.

\subsection{Vehicle Control}
Vehicle control is achieved through a path-following driver model integrated with the CARLA application programming interface (API), consisting of two subsystems: lateral and longitudinal control. Lateral control is governed by a Pure Pursuit Controller, which computes the angular velocity required to steer the vehicle toward a dynamically selected look-ahead point on the planned trajectory. The look-ahead distance is adaptively adjusted based on the available path, allowing the controller to respond effectively to both straight sections and sharp turns. Steering commands derived from this process are applied directly to the vehicle via the CARLA interface.

Longitudinal control is managed by a Proportional-Integral-Derivative (PID) controller. The system was configured with a conservative target speed of \textbf{0.5--1.0 m/s} for path following, successfully achieving a stable exploration speed of \textbf{1.5 m/s} during mapping. As shown in Fig.~\ref{PID}, the comparison between the published target speeds and the actual measured speeds demonstrates the corrective action of the PID controller in aligning the vehicle’s motion with the desired trajectory.

\begin{figure}[htbp]
\centerline{\includegraphics[scale=0.27]{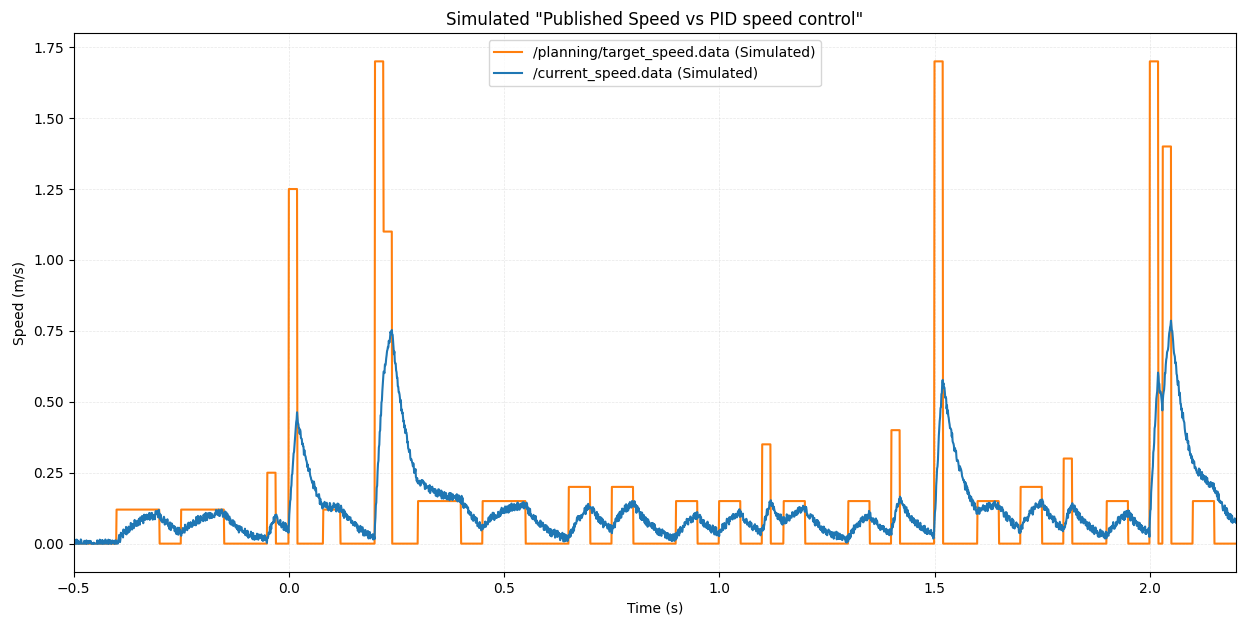}}
\caption{Correction published by the PID Controller}
\label{PID}
\end{figure}

\section{Conclusion}

The simulation results demonstrated stable cone detection, mapping, and trajectory execution, establishing a strong foundation for deployment on the physical ADS-DV platform equipped with a Jetson AGX Orin, ZED2i stereo camera, Robosense Helios 16P LiDAR, and CHCNAV INS. While the current validation has been limited to simulation, the modular framework ensures direct transferability to real-world testing.  

Future work will extend this pipeline to the physical ADS-DV, assess robustness under varied conditions, and benchmark computational performance. Further studies will compare fusion strategies and explore predictive controllers to optimize lap times. Overall, this highlights the potential of high-fidelity simulation with modular middleware for scalable autonomous racing development.

\bibliographystyle{ieeetr}
\bibliography{references/references}
\vspace{12pt}

\end{document}